%% file: main.tex
\documentclass{article}
\usepackage[dblblindworkshop,final]{neurips_2026}
\workshoptitle{Responsible Communication of Machine Learning Research in Biomedicine}
\usepackage[utf8]{inputenc}
\usepackage[T1]{fontenc}
\usepackage[hidelinks]{hyperref}
\usepackage{url}
\usepackage{booktabs}
\usepackage{amsfonts}
\usepackage{amsmath,amssymb}
\usepackage{nicefrac}
\usepackage{microtype}
\usepackage{xcolor}
\usepackage{graphicx}
\usepackage{caption}
\usepackage{subcaption}
\usepackage{placeins}
\usepackage{float}

\graphicspath{{figs/}}
\input{tables/macrosA}

\input{tables/defaultsA}
\title{Evaluation Choices Shape Biomedical ML Claims:\\A Pediatric Pneumonia Benchmark Case Study}
\author{%
  Bhanu Prakash Vangala \\
  University of Missouri \\
  \texttt{bv3hz@missouri.edu} \And
  Sowmya Guda \\
  University of Missouri \\
  \texttt{sghmy@missouri.edu} \AND
  Latha Peddi \\
  Government Degree College \\
  \texttt{lathakappala@outlook.com} \And
  Navya Vangala \\
  Amar Bio Tech Pvt Ltd \\
  \texttt{vangalanavya.8@gmail.com}
}
\begin{document}
\maketitle
\input{0_abstract}
\input{1_introduction}
\input{2_background}
\input{3_protocol}
\input{4_results}
\input{5_reporting}
\input{6_conclusion}
\bibliographystyle{plainnat}
\bibliography{refs}
\newpage
\appendix
\input{9_appendix}
\end{document}

%% file: tables/macrosA.tex
\newcommand{\AbestAccRaw}{0.858}

\newcommand{\AbestBacc}{0.811}

\newcommand{\AbestModel}{ViT-B/16}

\newcommand{\AdomAucAll}{0.697}
\newcommand{\AdomAucNormal}{0.898}
\newcommand{\AdomAucNull}{0.503}
\newcommand{\AdomAucPneu}{0.664}
\newcommand{\AdomAucTrainVal}{0.542}
\newcommand{\AmaxFtGain}{0.103}

\newcommand{\AmeanEceRaw}{0.172}
\newcommand{\AmeanEceValT}{0.169}
\newcommand{\AmeanFtGain}{0.044}
\newcommand{\AmeanOracleThreshGap}{0.083}
\newcommand{\AmeanThreshGain}{0.090}
\newcommand{\AmeanValT}{1.41}
\newcommand{\AmetaCV}{0.992}

\newcommand{\AmetaHoldout}{0.496}

\newcommand{\AnDupTest}{6}
\newcommand{\AnDupTrain}{26}
\newcommand{\AnEceTotal}{26}
\newcommand{\AnEceWorse}{9}

\newcommand{\AnTempBelowOne}{9}
\newcommand{\AnTest}{618}
\newcommand{\AnTrain}{4420}
\newcommand{\AnVal}{770}
\newcommand{\ApneuLeakAtNinetyFive}{14}

\newcommand{\AshapConfPneu}{0.997}
\newcommand{\AshapOutNormal}{17.0}
\newcommand{\AshapOutNormalSd}{7.7}
\newcommand{\AshapOutPneu}{34.5}
\newcommand{\AshapOutPneuSd}{10.3}

%% file: tables/defaultsA.tex
\providecommand{\AbestModel}{\textbf{??}}

\providecommand{\AdomAucAll}{\textbf{??}}
\providecommand{\AdomAucNormal}{\textbf{??}}
\providecommand{\AdomAucNull}{\textbf{??}}
\providecommand{\AdomAucPneu}{\textbf{??}}
\providecommand{\AdomAucTrainVal}{\textbf{??}}
\providecommand{\AmaxFtGain}{\textbf{??}}

\providecommand{\AmeanEceRaw}{\textbf{??}}
\providecommand{\AmeanEceValT}{\textbf{??}}
\providecommand{\AmeanFtGain}{\textbf{??}}
\providecommand{\AmeanOracleThreshGap}{\textbf{??}}
\providecommand{\AmeanThreshGain}{\textbf{??}}
\providecommand{\AmeanValT}{\textbf{??}}
\providecommand{\AmetaCV}{\textbf{??}}
\providecommand{\AmetaHoldout}{\textbf{??}}
\providecommand{\AnDupTest}{\textbf{??}}
\providecommand{\AnDupTrain}{\textbf{??}}
\providecommand{\AnEceTotal}{\textbf{??}}
\providecommand{\AnEceWorse}{\textbf{??}}

\providecommand{\AnTempBelowOne}{\textbf{??}}
\providecommand{\AnTest}{\textbf{??}}
\providecommand{\AnTrain}{\textbf{??}}
\providecommand{\AnVal}{\textbf{??}}
\providecommand{\ApneuLeakAtNinetyFive}{\textbf{??}}
\providecommand{\AshapConfPneu}{\textbf{??}}
\providecommand{\AshapOutNormal}{\textbf{??}}
\providecommand{\AshapOutNormalSd}{\textbf{??}}
\providecommand{\AshapOutPneu}{\textbf{??}}
\providecommand{\AshapOutPneuSd}{\textbf{??}}
\providecommand{\AassumedPrev}{\textbf{??}}

%% file: 0_abstract.tex
\begin{abstract}
Biomedical machine learning papers often compress model performance into one headline number. That number can look like a property of the model even when it depends strongly on how the benchmark was evaluated. We study this problem on the widely used Kermany pediatric chest radiograph dataset using nine image classifiers and a controlled evaluation protocol. Under the same protocol, the eight pretrained backbones differ by only 0.026 AUROC. In contrast, changing whether the backbone is frozen or fine tuned changes AUROC by \AmeanFtGain{} on average, and changing the decision threshold changes balanced accuracy by \AmeanThreshGain{} on average. The official test split is also measurably different from the training pool: a partition classifier distinguishes them at AUC \AdomAucAll{}, rising to \AdomAucNormal{} for normal radiographs. Most strikingly, a classifier using only file properties, with no image anatomy, reaches \AmetaCV{} balanced accuracy within the training pool but falls to \AmetaHoldout{} on the official test split. Validation fitted thresholds and calibration also transfer imperfectly. These results show that a high benchmark score can support different conclusions when the split, training policy, threshold, metric, calibration, and uncertainty are not communicated with it. We end with a seven item reporting recommendation in which each item is tied to an effect measured in the study.
\end{abstract}

%% file: 1_introduction.tex
\section{Introduction}

A biomedical machine learning paper is often remembered by one number. A reader may remember 98\% accuracy or 0.97 AUROC long after the details of the experiment are forgotten. That number can then appear in related work, presentations, grant proposals, or discussions of clinical readiness. The problem is that the number is usually read as a property of the model, even though it also depends on how the model was evaluated.

\paragraph{Problem statement.} Let $f_\theta$ denote a trained classifier, $\mathcal{D}$ the dataset, and $\pi$ the evaluation protocol. The reported score is
\begin{equation}
s = S\!\left(f_\theta,\pi;\mathcal{D}\right),
\label{eq:score}
\end{equation}
where $S$ is the scoring procedure. In this study, the protocol contains five choices that are common in biomedical image classification:
\begin{equation}
\pi = \left(\underbrace{\sigma}_{\text{split}},\underbrace{\phi}_{\text{backbone training}},\underbrace{t}_{\text{threshold}},\underbrace{T}_{\text{calibration}},\underbrace{m}_{\text{metric}}\right).
\label{eq:protocol}
\end{equation}
Papers usually describe the architecture $f_\theta$ in detail, but the elements of $\pi$ may be omitted, treated as defaults, or described only deep in the methods. If changing one element of $\pi$ changes the score as much as changing the architecture, then a headline score cannot be interpreted without the protocol that produced it.

We make that comparison explicit. For a fixed reference protocol $\pi_0$ and a set of architectures $\mathcal{F}$, we define the architectural spread and the effect of changing one protocol component $j$ as
\begin{equation}
\Delta_{\mathrm{arch}} = \max_{f\in\mathcal{F}} S(f,\pi_0;\mathcal{D})-\min_{f\in\mathcal{F}} S(f,\pi_0;\mathcal{D}), \qquad \Delta_{\pi}^{(j)} = \left|S(f,\pi;\mathcal{D})-S(f,\pi';\mathcal{D})\right|,
\label{eq:deltas}
\end{equation}
where $\pi$ and $\pi'$ differ only in component $j$. Equation~\ref{eq:deltas} gives the paper a simple question: are differences attributed to models larger than differences created by ordinary evaluation choices?

\paragraph{Case study.} We study the pediatric chest radiograph dataset released by Kermany et al.~\citep{kermany2018}. It is a useful case because it is public, small enough for controlled experiments, widely reused, and distributed with a fixed train and test split. Published studies report high performance on this corpus, often under evaluation setups that are not directly comparable~\citep{stephen2019pneumonia,kundu2021ensemble,siddiqi2024survey}. Some preserve the official test split, while others pool the corpus and create a new split. Some freeze a pretrained backbone, while others fine tune it. Threshold selection and calibration are often not reported. These are exactly the choices represented by Eq.~\ref{eq:protocol}.

\paragraph{Main findings.} First, under one fixed protocol, the eight pretrained backbones differ by only 0.026 AUROC, while changing whether the backbone is frozen or fine tuned changes AUROC by \AmeanFtGain{} on average and as much as \AmaxFtGain{}. Second, the official test split is measurably different from the training pool. A partition classifier distinguishes the two at AUC \AdomAucAll{}, rising to \AdomAucNormal{} for normal radiographs. Third, a model that uses only file properties and no anatomy reaches \AmetaCV{} balanced accuracy within the training pool but only \AmetaHoldout{} on the official test split. This makes the risk of pooling and randomly re-splitting the corpus concrete. Fourth, thresholds and calibration fitted on validation data do not fully transfer to the official test distribution. These results are summarized in Table~\ref{tab:effects} and developed in Section~\ref{sec:results}.

\paragraph{Contribution to responsible communication.} Our contribution is not a new pneumonia classifier and not a claim that this benchmark is unusable. It is a controlled case study of how the same biomedical benchmark can support different stories when the evaluation protocol is hidden. We connect each communication recommendation to a measured effect, rather than presenting a generic checklist. This directly addresses the gap between technical ML results and the claims that readers, clinicians, reviewers, and policymakers may take from them. Existing reporting guidance such as TRIPOD+AI already asks for many of these details~\citep{collins2024tripod}; our goal is to show, with measured examples, why those details change the meaning of the result.

%% file: 2_background.tex
\section{Background and Related Work}

\paragraph{Benchmark performance can reflect shortcuts rather than the intended signal.} High performance on a medical imaging dataset does not by itself show that a model learned clinically meaningful evidence. Models can exploit acquisition patterns, hospital identity, image processing, text markers, or other dataset specific features that correlate with the label but do not transfer to a new setting~\citep{zech2018variable,degrave2021shortcuts}. Shortcut learning is a broader machine learning problem in which a predictor uses an easy correlated feature instead of the mechanism the task is intended to measure~\citep{geirhos2020shortcut}. This distinction matters for communication because the headline metric describes predictive performance, not what evidence the model used.

\paragraph{Evaluation design can change the apparent conclusion.} Medical AI studies are especially sensitive to small test sets, leakage, data reuse, class imbalance, and unstable model rankings~\citep{varoquaux2022failures,roberts2021pitfalls}. Related benchmark studies show that duplicates can change rankings and that newly collected test sets can reveal drops that were hidden by the original evaluation~\citep{barz2020purging,recht2019imagenet}. These observations motivate our controlled comparison of architecture effects with protocol effects in Eq.~\ref{eq:deltas}. We also audit duplicates separately because contamination and distribution shift can produce superficially similar performance patterns.

\paragraph{Thresholds and calibration are part of the claim.} AUROC measures ranking over thresholds, but a decision system eventually operates at one threshold and presents confidence values that users may interpret as probabilities. Youden's index is a standard way to choose an operating point when sensitivity and specificity are weighted equally~\citep{youden1950}. Temperature scaling is a standard post hoc calibration method that often improves confidence estimates on data drawn from the same distribution as the validation set~\citep{guo2017calibration}. When validation and test distributions differ, however, the chosen threshold and calibration map may not transfer. This is important in medical AI because a model can preserve ranking performance while producing poorly chosen decisions or misleading confidence values~\citep{liu2022audit}.

\paragraph{Reporting standards address many of these details, but the communication gap remains.} TRIPOD+AI asks authors to describe evaluation data, discrimination, calibration, thresholds, and uncertainty for prediction model studies~\citep{collins2024tripod}. The challenge is not only whether such information exists somewhere in a paper. The challenge is whether the performance claim that travels outside the paper retains enough context to be interpreted correctly. Our study therefore treats the benchmark score itself as a communication object and measures how much meaning is lost when the protocol around that score is omitted.

%% file: 3_protocol.tex
\section{Study Design}
\label{sec:protocol}

This section defines the quantities used in the Results so that each reported effect corresponds to a specific part of the protocol in Eq.~\ref{eq:protocol}. Let $p_i\in[0,1]$ be the predicted probability of pneumonia for case $i$, $y_i\in\{0,1\}$ its label, and $z_i$ the model logits.

\paragraph{Dataset and split policy $\sigma$.}\label{sec:split} The Kermany corpus contains pediatric chest radiographs labeled \textsc{normal} or \textsc{pneumonia}, distributed as 5,216 training images, 16 validation images, and 624 test images~\citep{kermany2018}. The official validation set contains only eight images per class, so it is too small for stable early stopping, threshold selection, or calibration. We therefore create a patient disjoint validation split from the released training pool and leave the official test split untouched. Filenames encode a patient or study identifier, so related images are assigned as a group. We remove exact byte level duplicates before training, yielding \AnTrain{} training images, \AnVal{} validation images, and \AnTest{} official test images. Figure~\ref{fig:data} in Appendix~\ref{app:data} shows the split composition. A separate near duplicate audit in the same appendix checks that train to test leakage is not the explanation for the distributional effects reported in Section~\ref{sec:shift}.

\paragraph{Models and backbone training $\phi$.} We evaluate a small CNN trained from scratch and eight pretrained backbones: ResNet-18 and ResNet-50~\citep{he2016resnet}, VGG-16~\citep{simonyan2015vgg}, EfficientNet-B0~\citep{tan2019efficientnet}, DenseNet-121~\citep{huang2017densenet}, ConvNeXt-T~\citep{liu2022convnext}, ViT-B/16~\citep{dosovitskiy2021vit}, and Swin-T~\citep{liu2021swin}. For the main comparison, all pretrained backbones are fully fine tuned using the same split, augmentation, optimizer family, training budget, and three random seeds. To isolate $\Delta_{\pi}^{(\phi)}$ in Eq.~\ref{eq:deltas}, we repeat each pretrained architecture with the backbone frozen and only the classifier head trainable. Full per architecture results are provided in Appendix~\ref{app:fullmetrics}.

\paragraph{Decision threshold $t$ and metric $m$.} A binary prediction is $\hat y_i=\mathbb{1}[p_i\ge t]$. The official test set is 62.5\% positive, so an always positive classifier already obtains 0.625 raw accuracy. We therefore emphasize balanced accuracy for threshold dependent comparisons and report AUROC and AUPRC for ranking performance. We define
\begin{equation}
\mathrm{BA}(t)=\frac{\mathrm{TPR}(t)+\mathrm{TNR}(t)}{2}, \qquad t^{\star}=\operatorname*{arg\,max}_{t}\left[\mathrm{TPR}_{\mathrm{val}}(t)+\mathrm{TNR}_{\mathrm{val}}(t)-1\right].
\label{eq:ba}
\end{equation}
The first quantity is the reported balanced accuracy and the second is the validation selected threshold using Youden's $J$~\citep{youden1950}. Section~\ref{sec:posthoc} compares this threshold with the default $t=0.5$ and with a test selected oracle used only to measure the transfer gap.

\paragraph{Calibration $T$.} We fit one temperature on validation logits by minimizing cross entropy and measure calibration using expected calibration error with 15 equal width confidence bins~\citep{guo2017calibration}:
\begin{equation}
T^{\star}=\operatorname*{arg\,min}_{T>0}-\sum_{i\in\mathrm{val}}\log\operatorname{softmax}(z_i/T)_{y_i}, \qquad \mathrm{ECE}=\sum_{b=1}^{B}\frac{|b|}{N}\left|\operatorname{acc}(b)-\operatorname{conf}(b)\right|.
\label{eq:temp}
\end{equation}
A value $T^{\star}<1$ sharpens the probabilities and $T^{\star}>1$ softens them. Section~\ref{sec:posthoc} evaluates whether the validation fitted correction improves calibration on the official test split, and Appendix~\ref{app:calib} provides the full run level results and reliability diagram.

\paragraph{Distribution shift test.} To test whether two partitions are exchangeable, we extract frozen ResNet-50 features and train logistic regression to predict whether an image came from the training pool or the official test split. We use five fold out of fold predictions and repeat the analysis after shuffling partition labels. In addition to discriminator AUC, we report the proxy $\mathcal{A}$ distance~\citep{bendavid2010}:
\begin{equation}
d_{\mathcal{A}}=2(1-2\epsilon), \qquad \epsilon=\text{balanced error of the partition classifier}.
\label{eq:adist}
\end{equation}
A value near 0 means that the partitions are difficult to distinguish, while a value near 2 means that they are easily separable. We repeat the procedure on our own train and validation partitions as a control.

\paragraph{File only shortcut test.} We train a gradient boosted tree using only file properties: pixel dimensions, aspect ratio, file size, bytes per pixel, image mode, and JPEG quantization tables. It receives no image pixels and no anatomical features. We first evaluate it with five fold cross validation inside the training pool, then fit it on that pool and evaluate it on the untouched official test split. A large drop between these two evaluations would show that file construction artifacts encode the label within one partition but do not transfer to the official test set.

\paragraph{Target prevalence operating point.} The test selected threshold in Section~\ref{sec:posthoc} is only an oracle and cannot be used in a valid evaluation. To test whether target information can recover some of that gap without labels, we also consider an operating point that matches an assumed target prevalence $\hat\pi$:
\begin{equation}
t_{\hat\pi}=\inf\left\{t:\frac{1}{n}\sum_{i=1}^{n}\mathbb{1}[p_i\ge t]\le\hat\pi\right\}.
\label{eq:prior}
\end{equation}
This method uses unlabeled target predictions plus an externally specified prevalence. We use $\hat\pi=\AassumedPrev{}$ only as a sensitivity analysis, with the result and its limitations reported in Section~\ref{sec:posthoc} and Appendix~\ref{app:calib}.

\paragraph{Uncertainty and statistical testing.} We report mean and standard deviation over three seeds where available. Confidence intervals for AUROC are obtained by percentile bootstrap resampling of test cases with 2,000 resamples, and paired AUROC differences are tested using DeLong's method~\citep{delong1988}. Precision-recall curves and AUPRC are also reported because class imbalance can make precision-recall summaries more informative than ROC summaries for some comparisons~\citep{saito2015prc}. Statistical details and the full curves are in Appendices~\ref{app:stats} and~\ref{app:curves}.

%% file: 4_results.tex
\section{Results}
\label{sec:results}

Table~\ref{tab:a-main} gives the main model results on the untouched official test split, and Table~\ref{tab:effects} summarizes the effects associated with the protocol components in Eq.~\ref{eq:protocol}. The full metric tables are moved to Appendix~\ref{app:fullmetrics} so that the main text can focus on the comparisons that change the interpretation of the benchmark score.

\input{tables/tabA_main_short}
\input{tables/tabA_effects}

\subsection{Architecture differences are small relative to several protocol effects}
\label{sec:arch}

Under the reference protocol $\pi_0$, the eight pretrained architectures occupy a narrow AUROC range of 0.026. Their bootstrap intervals overlap substantially in Figure~\ref{fig:pair}a, and the paired DeLong analysis in Appendix~\ref{app:stats} shows that most pairwise differences are not distinguishable at this test set size. The custom CNN trained from scratch is clearly weaker, but the ordering among the pretrained backbones should not be read as a stable leaderboard.

The uncertainty is also large relative to the architectural spread. With \AnTest{} test images after exact duplicate removal, the 95\% bootstrap interval is roughly $\pm0.02$ AUROC for individual models. This means that reporting a score to three decimal places without an interval can communicate more precision than the benchmark supports. In the notation of Eq.~\ref{eq:deltas}, $\Delta_{\mathrm{arch}}=0.026$ is the reference against which we compare the protocol effects below.

\begin{figure}[t]
\centering
\begin{subfigure}[b]{0.40\linewidth}
\centering
\includegraphics[width=\linewidth]{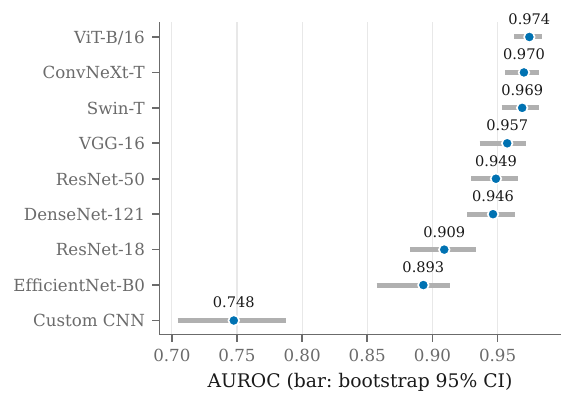}
\caption{AUROC with bootstrap 95\% confidence intervals.}
\end{subfigure}\hfill
\begin{subfigure}[b]{0.40\linewidth}
\centering
\includegraphics[width=\linewidth]{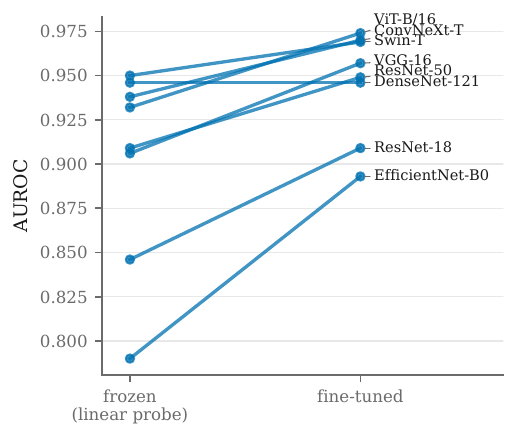}
\caption{Frozen and fine tuned backbones.}
\end{subfigure}
\caption{Two views of protocol sensitivity. Panel (a) shows that the pretrained architectures have heavily overlapping uncertainty intervals under one fixed protocol. Panel (b) holds architecture, split, augmentation, and training schedule fixed while changing only whether the pretrained backbone is frozen or fine tuned. The crossing lines show that the architecture ranking is not preserved.}
\label{fig:pair}
\end{figure}

\subsection{Backbone training changes the conclusion more than architecture choice}
\label{sec:frozen}

Changing $\phi$ from a frozen feature extractor to full fine tuning changes AUROC by \AmeanFtGain{} on average across the eight pretrained architectures and by as much as \AmaxFtGain{}. Both values are larger than the 0.026 architectural spread defined above. Figure~\ref{fig:pair}b also shows that the sign of the change is not constant. Some models improve after fine tuning, while others change little or lose balanced accuracy, so a ranking obtained under one backbone training policy cannot simply be transferred to the other.

This comparison is intentionally simple because the communication failure is simple. Two papers can name the same architecture but train it in materially different ways. If the backbone training policy is absent from the headline comparison, a reader may attribute a difference to architecture design when it is partly or mainly a protocol effect. Appendix Table~\ref{tab:a-frozen} provides the per architecture values for balanced accuracy, $F_1$, and AUROC.

\subsection{The official test split is measurably different from the training pool}
\label{sec:shift}

\begin{figure}[H]
\centering
\begin{subfigure}[b]{0.57\linewidth}
\centering
\includegraphics[width=\linewidth]{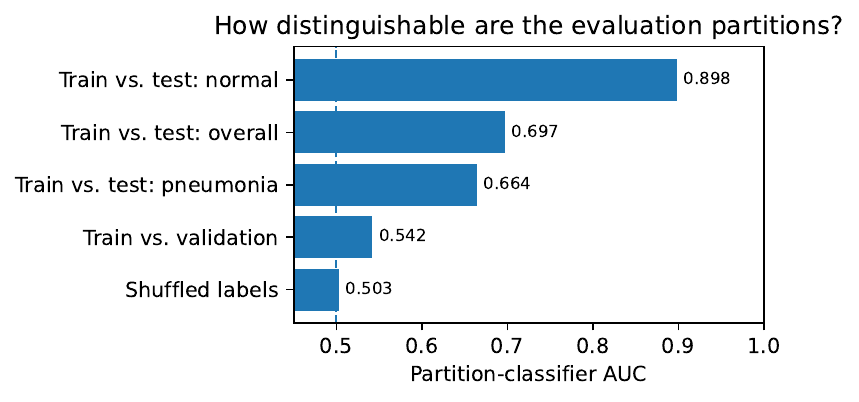}
\caption{Partition distinguishability. The dashed line marks chance AUC.}
\end{subfigure}\hfill
\begin{subfigure}[b]{0.38\linewidth}
\centering
\includegraphics[width=\linewidth]{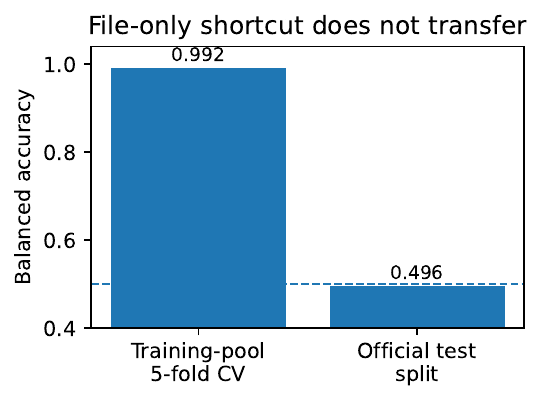}
\caption{File only label prediction across split policies.}
\end{subfigure}
\caption{Evidence that split policy changes what the benchmark measures. Panel (a) shows that the official test split is distinguishable from the training pool, especially for normal radiographs, while the study's own train and validation partitions are much closer to chance. Panel (b) shows that file properties alone nearly solve the label task inside the training pool but fail on the untouched official test split. Together, these results explain why pooling and randomly re-splitting the released data can preserve a shortcut that does not transfer.}
\label{fig:split-evidence}
\end{figure}

The partition classifier separates the training pool from the official test split at AUC \AdomAucAll{}, compared with \AdomAucNull{} after shuffling partition labels. The shift is not uniform across classes. For \textsc{normal} radiographs, partition AUC rises to \AdomAucNormal{}, while for \textsc{pneumonia} it is \AdomAucPneu{}. Applying Eq.~\ref{eq:adist} gives a proxy $\mathcal{A}$ distance of 1.226 for the normal class, compared with 0.117 for our own patient disjoint train and validation split. Figure~\ref{fig:split-evidence}a makes this contrast visible, and Appendix Table~\ref{tab:a-transfer-bc} gives the corresponding proxy $\mathcal{A}$ distances.

Two controls make the interpretation more specific. First, the same partition classifier applied to our train and validation partitions reaches only \AdomAucTrainVal{}, which is close to chance. Second, the duplicate audit in Appendix~\ref{app:data} finds no test image with a near twin in the training pool at the primary similarity threshold. These checks make simple duplication an unlikely explanation. The official test split therefore behaves as a small transfer set rather than as an exchangeable sample from the training pool.

We cannot identify the source of the shift because the released metadata do not contain the acquisition and curation information needed to do so. Similar hidden acquisition and site effects are well documented in medical imaging datasets~\citep{zech2018variable,oakdenrayner2020exploring}. Our claim is therefore limited to what the experiment establishes: the partitions are distinguishable, the difference is strongest for normal images, and post hoc choices fitted on the training distribution should not be assumed to transfer unchanged.

\subsection{A file only predictor shows why pooling and re-splitting can be misleading}
\label{sec:meta}

The file only classifier makes the split problem easy to see. It uses image dimensions, file size, bytes per pixel, image mode, and JPEG quantization information, but no radiograph pixels and therefore no anatomy. Within the training pool it predicts the diagnosis at \AmetaCV{} balanced accuracy under five fold cross validation. When fitted on that pool and evaluated on the untouched official test split, balanced accuracy falls to \AmetaHoldout{}, which is essentially chance. Figure~\ref{fig:split-evidence}b shows this collapse next to the partition shift result so that the two pieces of evidence can be read together.

The result should not be interpreted as evidence that file headers cause pneumonia. It shows that the two labels in the training pool were stored or processed in systematically different ways. Those construction artifacts provide an easy shortcut inside that pool, but the shortcut does not survive the move to the official test split. This pattern is consistent with the broader shortcut learning literature, where nonclinical acquisition signals can support apparently strong within dataset performance~\citep{degrave2021shortcuts,geirhos2020shortcut}.

This directly explains why split policy $\sigma$ is part of the reported score in Eq.~\ref{eq:score}. If the released partitions are pooled and randomly re-split, train and test examples can share the same file level signature. The resulting evaluation may reward regularities that disappear on the official split. The pair \AmetaCV{} within the training pool and \AmetaHoldout{} on the official test set provides a concrete measured example of how a seemingly small evaluation decision can change the story attached to a high score.

\subsection{Thresholds and calibration fitted on validation do not fully transfer}
\label{sec:posthoc}

The decision threshold is another large protocol effect. Moving from the default $t=0.5$ to the validation selected $t^{\star}$ in Eq.~\ref{eq:ba} changes balanced accuracy by \AmeanThreshGain{} on average, which is larger than $\Delta_{\mathrm{arch}}$. The full per architecture comparison is in Appendix Table~\ref{tab:a-calib-full}. Figure~\ref{fig:posthoc}a gives the case count view for the strongest model and shows why the threshold must be reported together with any label based metric.

Validation tuning does not remove the entire operating point problem. Choosing the threshold directly on the test labels, which is an invalid oracle used only to measure the gap, improves balanced accuracy by a further \AmeanOracleThreshGap{} on average. The target prevalence rule in Eq.~\ref{eq:prior} is a sensitivity analysis that does not use target labels. In the aggregate experiment it raises balanced accuracy from 0.787 to 0.863, compared with the unattainable test oracle of 0.870, recovering 92\% of that specific gap. Appendix Table~\ref{tab:a-transfer} reports these values. We do not propose this as a general deployment rule because it depends on a credible external prevalence estimate. We include it to show that the failure is connected to the target distribution rather than to a lack of threshold optimization.

Calibration shows a related transfer problem. Using Eq.~\ref{eq:temp}, mean test ECE changes only from \AmeanEceRaw{} before calibration to \AmeanEceValT{} after applying a temperature fitted on validation data, and ECE becomes worse in \AnEceWorse{} of \AnEceTotal{} runs. The mean fitted temperature is \AmeanValT{}, and $T^{\star}<1$ occurs in \AnTempBelowOne{} runs, so the direction of the fitted correction is not consistent across models. Figure~\ref{fig:posthoc}b shows the corresponding reliability diagram for \AbestModel{}, while Appendix Table~\ref{tab:a-transfer-bc} provides the Brier decomposition. The conclusion is not that temperature scaling is ineffective in general. It is that the phrase ``the model was calibrated'' is incomplete unless the paper states where calibration was fitted and where it was evaluated.

\begin{figure}[H]
\centering
\begin{subfigure}[b]{0.48\linewidth}
\centering
\includegraphics[width=\linewidth]{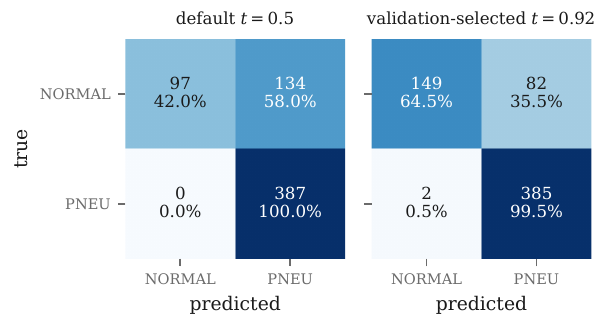}
\caption{Default versus validation selected threshold for \AbestModel{}.}
\end{subfigure}\hfill
\begin{subfigure}[b]{0.48\linewidth}
\centering
\includegraphics[width=\linewidth]{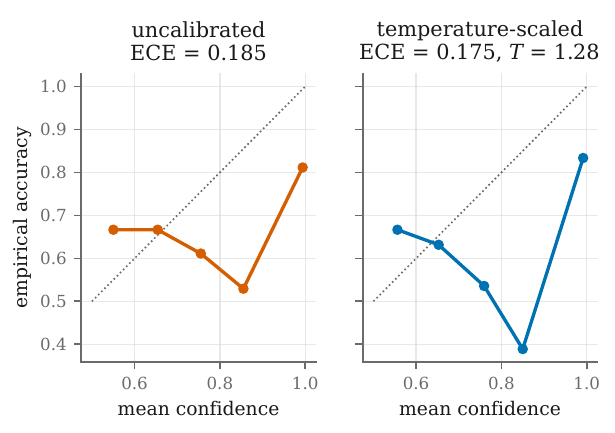}
\caption{Reliability before and after validation fitted temperature scaling.}
\end{subfigure}
\caption{Post hoc choices that change the reported interpretation on the official test split. Panel (a) shows how threshold selection changes the false positive and true negative counts while leaving the model itself unchanged. Panel (b) shows that a temperature fitted on validation data changes confidence estimates but does not remove the calibration gap on the official test distribution.}
\label{fig:posthoc}
\end{figure}

%% file: tables/tabA_main_short.tex
\begin{table}[t]
\centering
\caption{Performance on the untouched official test split (mean$\pm$s.d. over three seeds). All pretrained backbones are fully fine-tuned, and the decision threshold is selected on the patient-disjoint validation split. The pretrained models have similar AUROC despite different architectures.}
\label{tab:a-main}
\setlength{\tabcolsep}{4pt}
\scriptsize
\begin{tabular}{l rrrrr}
\toprule
Architecture & Bal. Acc. & Sens. & Spec. & AUROC & ECE $\downarrow$ \\
\midrule
Custom CNN (scratch) & 0.697$\pm$0.032 & 0.881$\pm$0.027 & 0.512$\pm$0.076 & 0.748$\pm$0.042 & 0.178$\pm$0.044 \\
\midrule
ResNet-18 & 0.768$\pm$0.016 & 0.967$\pm$0.019 & 0.569$\pm$0.038 & 0.909$\pm$0.006 & 0.143$\pm$0.027 \\
VGG-16 & 0.801$\pm$0.023 & 0.996$\pm$0.001 & 0.606$\pm$0.047 & 0.957$\pm$0.007 & 0.150$\pm$0.018 \\
ResNet-50 & \textbf{0.828$\pm$0.006} & 0.984$\pm$0.000 & \textbf{0.671$\pm$0.013} & 0.949$\pm$0.005 & 0.163$\pm$0.005 \\
EfficientNet-B0 & 0.784$\pm$0.011 & 0.963$\pm$0.007 & 0.605$\pm$0.029 & 0.893$\pm$0.008 & 0.191$\pm$0.001 \\
DenseNet-121 & 0.775$\pm$0.041 & 0.992$\pm$0.006 & 0.557$\pm$0.084 & 0.946$\pm$0.014 & 0.194$\pm$0.008 \\
ConvNeXt-T & 0.820$\pm$0.035 & 0.994$\pm$0.003 & 0.646$\pm$0.071 & 0.970$\pm$0.008 & 0.181$\pm$0.016 \\
ViT-B/16 & 0.811$\pm$0.022 & 0.996$\pm$0.001 & 0.626$\pm$0.043 & \textbf{0.974$\pm$0.008} & 0.143$\pm$0.048 \\
Swin-T & 0.809$\pm$0.016 & 0.995$\pm$0.000 & 0.623$\pm$0.031 & 0.969$\pm$0.005 & 0.177$\pm$0.028 \\
\bottomrule
\end{tabular}
\vspace{1pt}\\[-3pt]\parbox{\linewidth}{\scriptsize ECE: expected calibration error. Sensitivity is for the \textsc{pneumonia} class. Raw accuracy is not emphasized because the official test set is 62.5\% positive.}
\end{table}

%% file: tables/tabA_effects.tex
\begin{table}[t]
\centering
\caption{Observed effect of the evaluation choices defined in Eq.~\ref{eq:protocol}. The architecture row gives the AUROC range across the eight pretrained backbones under one fixed protocol. Other rows change one evaluation component or test whether the evaluation partitions differ. Oracle values are used only to measure transfer gaps.}
\label{tab:effects}
\setlength{\tabcolsep}{4pt}
\footnotesize
\begin{tabular}{p{0.29\linewidth} p{0.31\linewidth} p{0.30\linewidth}}
\toprule
Question & Comparison & Observed effect \\
\midrule
Architecture under fixed $\pi_0$ & eight pretrained backbones & AUROC range 0.026 \\
Backbone training $\phi$ & frozen vs. fine tuned & mean $|\Delta|$ \AmeanFtGain{} AUROC; max \AmaxFtGain{} \\
Threshold $t$ & 0.5 vs. validation selected & mean $|\Delta|$ \AmeanThreshGain{} balanced accuracy \\
Threshold transfer & validation vs. test oracle & mean gap \AmeanOracleThreshGap{} balanced accuracy \\
Calibration $T$ & raw vs. validation fitted & ECE \AmeanEceRaw{} to \AmeanEceValT{}; worse in \AnEceWorse{}/\AnEceTotal{} runs \\
Split $\sigma$ and file shortcut & training pool CV vs. official test & \AmetaCV{} to \AmetaHoldout{} balanced accuracy \\
Train to test shift & partition classifier & AUC \AdomAucAll{} overall; \AdomAucNormal{} on normal images \\
\bottomrule
\end{tabular}
\end{table}

%% file: 5_reporting.tex
\section{From a Benchmark Score to a Responsible Claim}
\label{sec:reporting}

\paragraph{Published numbers on the same corpus are not automatically comparable.} Table~\ref{tab:a-lit} places two published results beside our evaluation to show how different protocols can sit behind similar headline numbers. Stephen et al.~\citep{stephen2019pneumonia} report 0.937 accuracy after pooling and re-splitting the data, while Kundu et al.~\citep{kundu2021ensemble} report 0.988 accuracy under pooled cross validation. Our rows preserve the official test split. Because split policy, metric, model training, and operating point differ, these values answer different questions rather than forming one valid ranking. Equation~\ref{eq:protocol} makes those differences explicit.

\input{tables/tabA_lit}

\paragraph{Seven details should travel with the headline score.} The measured effects in Table~\ref{tab:effects} suggest a compact reporting rule. A biomedical benchmark result should state the evaluation split, whether pretrained features were frozen or fine tuned, the decision threshold and how it was selected, a metric that exposes class imbalance, how and where calibration was fitted, uncertainty or an appropriate paired comparison, and a basic train to test shift check when practical. These are not stylistic preferences. Each item corresponds to an effect measured in Sections~\ref{sec:arch} to~\ref{sec:posthoc}.

This recommendation is consistent with the broader goals of TRIPOD+AI~\citep{collins2024tripod}, but our point is narrower: the cost of missing protocol details can be measured on the benchmark itself. Here, backbone training changes AUROC more than the pretrained architecture spread, threshold selection changes balanced accuracy by even more, the file only shortcut falls from \AmetaCV{} to \AmetaHoldout{} across split policies, and the official test partition is measurably different from the training pool.

\paragraph{A benchmark metric cannot say what evidence the model used.} Appendix~\ref{app:attr} provides a supplementary Grad-CAM~\citep{selvaraju2017gradcam} and SHAP~\citep{lundberg2017shap} audit. We summarize attribution mass outside a body mask rather than treating a heatmap as a causal explanation. This analysis is separate from the protocol results, but it illustrates the same communication limit: a strong discrimination score does not reveal whether the visual evidence used by a model is clinically meaningful.

\paragraph{What can safely be claimed from this benchmark?} The strongest supported claim is that modern pretrained image classifiers separate the two released labels well on the official split under a stated protocol. The experiments do not establish cross hospital generalization, clinical utility, readiness for triage, or meaningful superiority of one closely clustered pretrained architecture over another. Because the official split also contains an undocumented distribution difference, a responsible summary should report the score together with the split, metric, threshold, uncertainty, and observed shift.

\paragraph{Why this belongs in a communication workshop.} The failure occurs when a precise result is shortened into a claim that drops the protocol needed to interpret it. Table~\ref{tab:effects} shows the remedy: keep those technical conditions attached to the result by expressing them as measured changes in the reported score.

%% file: tables/tabA_lit.tex
\begin{table}[htbp]
\centering
\caption{Examples of reported results on the same dataset under different evaluation setups. These rows should not be read as a model ranking: the split and metric differ across studies. ``Pooled'' means that images were combined and re-partitioned rather than evaluated on the untouched official test split.}
\label{tab:a-lit}
\setlength{\tabcolsep}{3pt}
\footnotesize
\begin{tabular}{l l l r}
\toprule
Study & Method & Evaluation setup & Reported score \\
\midrule
\citep{stephen2019pneumonia} & CNN from scratch & pooled, re-split & 0.937 acc. \\
\citep{kundu2021ensemble} & 3-model ensemble & pooled, 5-fold CV & 0.988 acc. \\
\midrule
\multicolumn{4}{l}{\emph{This work: untouched official test split; threshold selected on validation}} \\
\phantom{xx}\AbestModel{} & fine-tuned & official test & \AbestAccRaw{} acc. \\
\phantom{xx}\AbestModel{} & fine-tuned & official test & \AbestBacc{} bal. acc. \\
\bottomrule
\end{tabular}
\end{table}

%% file: 6_conclusion.tex
\section{Limitations and Conclusion}

\paragraph{Limitations.} This is a case study of one public pediatric radiograph dataset, not a survey of biomedical benchmarks. The released metadata do not identify the source of the train to test shift, and the file only experiment establishes a construction shortcut without proving that a particular neural network uses it. The small test set limits architecture comparisons, while the target prevalence threshold is only a sensitivity analysis because reliable prevalence may not be available at deployment. Additional robustness and reproducibility details appear in Appendices~\ref{app:ensemble} and~\ref{app:repro}.

\paragraph{Conclusion.} A benchmark score reflects both model and protocol. Here, training, split, threshold, calibration, and shift choices affect the result as much as or more than architecture choice. Those conditions need to travel with the score.

%% file: 9_appendix.tex
\section*{Appendix Guide}

Appendix~\ref{app:fullmetrics} provides complete per model tables for Sections~\ref{sec:arch}, \ref{sec:frozen}, and~\ref{sec:posthoc}. Appendix~\ref{app:calib} provides transfer diagnostics for Sections~\ref{sec:posthoc} and~\ref{sec:shift}. Appendices~\ref{app:stats} and~\ref{app:data} document statistical testing and leakage controls. Appendix~\ref{app:curves} provides the underlying discrimination and error plots, Appendix~\ref{app:attr} contains the separate attribution diagnostic discussed in Section~\ref{sec:reporting}, Appendix~\ref{app:ensemble} gives a seed ensemble robustness check, and Appendix~\ref{app:repro} describes reproducibility artifacts.

\section{Full Model and Protocol Results}
\label{app:fullmetrics}

The main text reports only the columns needed for the central comparisons. Table~\ref{tab:a-main-full} gives the complete fine tuned model results, including $F_1$ and AUPRC, so that the shorter Table~\ref{tab:a-main} can be checked without repeating its discussion. Table~\ref{tab:a-frozen} gives the corresponding frozen versus fine tuned comparison for each pretrained architecture. Table~\ref{tab:a-calib-full} reports the per architecture effect of the default and validation selected thresholds together with ECE before and after temperature scaling. These tables support Sections~\ref{sec:arch}, \ref{sec:frozen}, and~\ref{sec:posthoc}; the interpretation remains in the main Results section.

\input{tables/tabA_main_full}
\input{tables/tabA_frozen_full}
\input{tables/tabA_calib_full}

\FloatBarrier
\section{Calibration and Operating Point Transfer}
\label{app:calib}

Section~\ref{sec:posthoc} reports the aggregate transfer effects and Figure~\ref{fig:posthoc} shows the threshold and calibration behavior that directly supports those claims. This appendix therefore keeps only the additional numerical decomposition. Table~\ref{tab:a-transfer} reports the validation threshold, target prevalence threshold from Eq.~\ref{eq:prior}, and the unattainable test oracle. Table~\ref{tab:a-transfer-bc} separates calibration reliability from resolution and reports the proxy $\mathcal{A}$ distances used in Section~\ref{sec:shift}. These tables provide values that are useful for verification but are not needed to repeat the main visual result.

\input{tables/tabA_transfer_a}
\input{tables/tabA_transfer_bc}

\FloatBarrier
\section{Statistical Testing of Architecture Differences}
\label{app:stats}

\begin{figure}[htbp]
\centering
\includegraphics[width=0.62\linewidth]{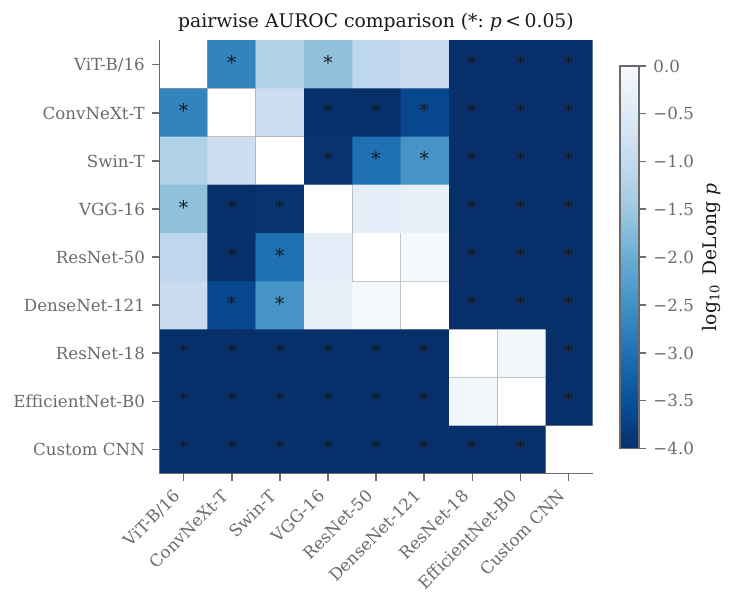}
\caption{Pairwise DeLong tests for AUROC on seed 0. Cells marked $*$ indicate $p<0.05$, and shading represents $\log_{10}p$. The figure supports the uncertainty discussion in Section~\ref{sec:arch}.}
\label{fig:delong}
\end{figure}

The overlapping confidence intervals in Figure~\ref{fig:pair} show uncertainty for each model separately. Figure~\ref{fig:delong} complements that view with paired DeLong tests on seed 0, which use the fact that the models are evaluated on the same test cases. The purpose of this analysis is to check the architecture comparison in Section~\ref{sec:arch}, not to create a second model ranking. At the available test set size, most differences among the pretrained backbones are not statistically distinguishable.

\FloatBarrier
\section{Corpus Composition and Duplicate Audit}
\label{app:data}

\begin{figure}[htbp]
\centering
\includegraphics[width=0.92\linewidth]{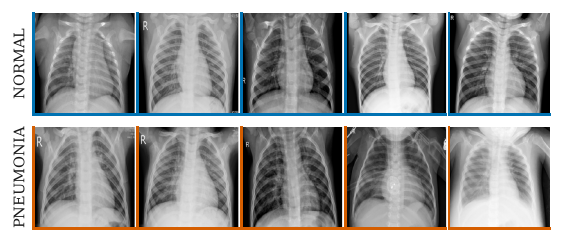}
\caption{Left: representative official test images from each class. Right: split composition after patient disjoint partitioning and exact duplicate removal. The official test set remains untouched apart from removal of exact duplicates.}
\label{fig:data}
\end{figure}

Figure~\ref{fig:data} shows representative test images and the class composition of the patient disjoint partitions used in this study. The figure also makes the class imbalance visible, which motivates balanced accuracy in Eq.~\ref{eq:ba}. The duplicate audit is reported separately from the shift analysis because a duplicated image and a shifted distribution can both produce unusual train to test behavior but have different explanations.

\paragraph{Near duplicate criterion.} Each image is converted to a $64\times64$ gray scale array $x$, mean centered to $\tilde{x}=x-\bar{x}\mathbf{1}$, and $L_2$ normalized. Similarity between two images is measured using zero mean normalized cross correlation,
\begin{equation}
\rho(x,y)=\frac{\langle\tilde{x},\tilde{y}\rangle}{\lVert\tilde{x}\rVert_2\lVert\tilde{y}\rVert_2}.
\label{eq:ncc}
\end{equation}
Equation~\ref{eq:ncc} is used only for the leakage audit. We use $\rho\ge0.98$ as the primary near duplicate threshold. At that threshold no official test image has a near twin in the training pool. At the looser threshold $\rho\ge0.95$, \ApneuLeakAtNinetyFive{} of 624 released test images are flagged for manual inspection. Exact byte level duplicates number \AnDupTrain{} in the released training set and \AnDupTest{} in the released test set and are removed before model training or evaluation. The threshold sweep is reported so that this leakage check is not dependent on one unexplained operating point.

\paragraph{Why we did not use perceptual hashing as the primary audit.} A 64 bit difference hash at Hamming radius 5 flags approximately 93\% of the test set, which is inconsistent with the verified pixel level comparison above. Chest radiographs share a highly similar global layout, so a coarse gradient hash collides on many distinct images. This sensitivity check is included because a duplicate detector that is inappropriate for the domain could incorrectly turn a distribution shift finding into a leakage claim.

\FloatBarrier
\section{Discrimination Curves and Error Structure}
\label{app:curves}

\begin{figure}[htbp]
\centering
\begin{minipage}[b]{0.48\linewidth}
\centering
\includegraphics[width=\linewidth]{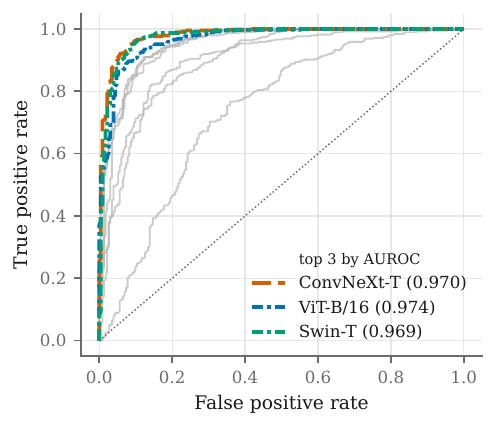}
\end{minipage}\hfill
\begin{minipage}[b]{0.48\linewidth}
\centering
\includegraphics[width=\linewidth]{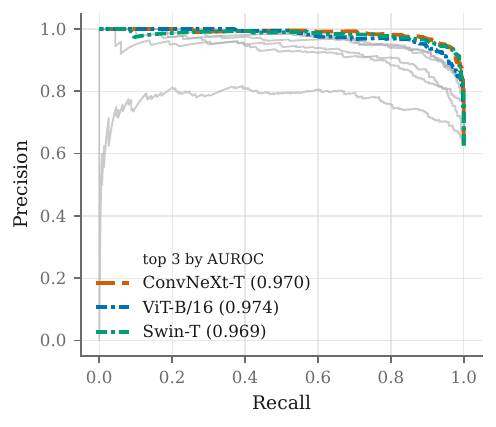}
\end{minipage}
\caption{ROC curve on the left and precision-recall curve on the right for seed 0. The three highest AUROC architectures are labeled and the remaining models are shown with lower visual emphasis. Precision-recall reporting follows the recommendation to inspect class imbalance sensitive summaries alongside ROC based summaries~\citep{saito2015prc}.}
\label{fig:rocpr}
\end{figure}

Table~\ref{tab:a-main-full} summarizes scalar performance, while Figure~\ref{fig:rocpr} shows the underlying ROC and precision-recall curves for seed 0. The threshold error structure is already shown in the main Results in Figure~\ref{fig:posthoc}a, so it is not repeated here. The curves below support the metric discussion in Section~\ref{sec:arch} without introducing a separate result claim.

\FloatBarrier
\section{Supplementary Attribution Audit}
\label{app:attr}

\begin{figure}[htbp]
\centering
\begin{minipage}[b]{0.52\linewidth}
\centering
\includegraphics[width=\linewidth]{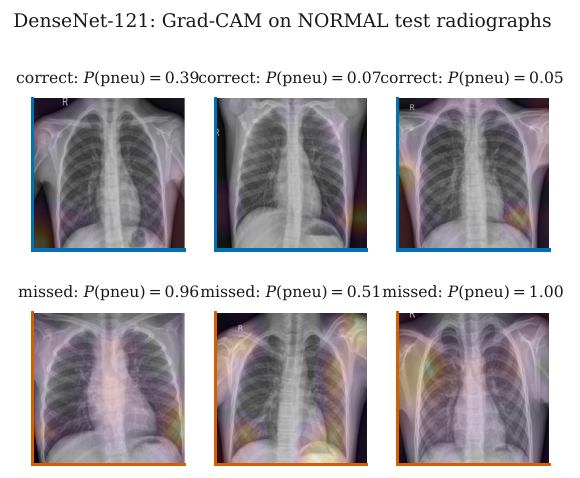}
\end{minipage}\hfill
\begin{minipage}[b]{0.44\linewidth}
\centering
\includegraphics[width=\linewidth]{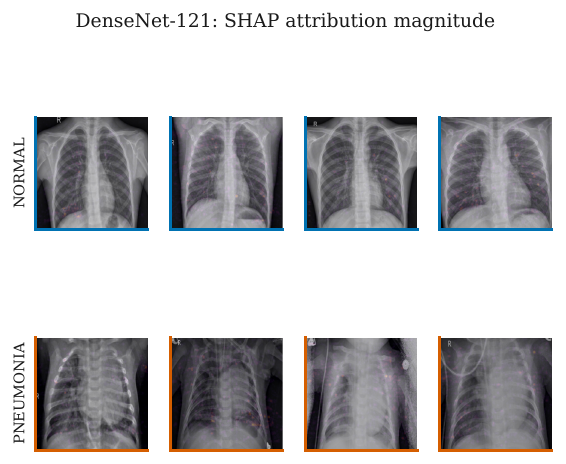}
\end{minipage}
\caption{Left: Grad-CAM examples on normal test radiographs. Right: SHAP attribution magnitude summarized by class. These visualizations are treated as descriptive diagnostics rather than as causal explanations of model reasoning.}
\label{fig:attr}
\end{figure}

The main paper does not use attribution as evidence for the protocol effects. This supplementary analysis instead illustrates the narrower communication point in Section~\ref{sec:reporting}: discrimination metrics do not reveal which image regions contribute to a prediction. Figure~\ref{fig:attr} shows Grad-CAM~\citep{selvaraju2017gradcam} examples and SHAP~\citep{lundberg2017shap} attribution summaries for a model retrained under the same configuration as the main benchmark runs.

To make the attribution analysis auditable, let $a_u$ denote the attribution assigned to pixel $u$ and let $\Omega$ be a body mask defined as the largest bright connected component after Otsu thresholding. We summarize the fraction of absolute attribution mass outside the body as
\begin{equation}
F_{\mathrm{out}}=\frac{\sum_{u\notin\Omega}|a_u|}{\sum_u|a_u|}.
\label{eq:fout}
\end{equation}
Equation~\ref{eq:fout} converts each attribution map to one scalar summary. For radiographs predicted as normal, the mean outside body fraction is \AshapOutNormal{}\% with standard deviation \AshapOutNormalSd{}\%. For radiographs predicted as pneumonia, it is \AshapOutPneu{}\% with standard deviation \AshapOutPneuSd{}\%, at mean confidence \AshapConfPneu{}. These values are not used to claim that the model is or is not clinically valid. They show why a performance number and a visual explanation answer different questions.

\FloatBarrier
\section{Seed Ensemble Robustness Check}
\label{app:ensemble}

Section~\ref{sec:arch} focuses on single model architecture comparisons because that is the communication question of interest. Table~\ref{tab:a-ensemble} provides a separate variance reduction check using only the three seeds already trained. Averaging their probabilities changes AUPRC modestly and does not alter the main conclusion that protocol effects can be larger than architecture differences.

\input{tables/tabA_ensemble}

\FloatBarrier
\section{Reproducibility}
\label{app:repro}

All reported numerical values are generated from stored per case model outputs into the LaTeX macro file used by the manuscript. The analysis stage consumes one metric record per run and one array of logits per run and split, which is sufficient to recompute metrics, operating points, confidence intervals, calibration analyses, and statistical tests without retraining. For an anonymized release, the intended artifacts are the patient disjoint split lists, per case predicted probabilities, the file only feature extractor used in Section~\ref{sec:meta}, the partition classifier used in Section~\ref{sec:shift}, and scripts that regenerate the tables and figures in this appendix. This organization keeps the numerical claims in the paper tied to reproducible analysis outputs rather than hand transcribed values.

%% file: tables/tabA_main_full.tex
\begin{table}[H]
\centering
\caption{Full version of Table~\ref{tab:a-main}: pneumonia detection on the held-out official test split (mean$\pm$s.d.\ over three seeds, all backbones fine-tuned). Operating point and temperature were selected on the patient-disjoint validation split only.}
\label{tab:a-main-full}
\setlength{\tabcolsep}{2.5pt}
\scriptsize
\begin{tabular}{l r rrrrrrr}
\toprule
Architecture & Par.\,(M) & Bal.\,Acc. & Sens. & Spec. & F1 & AUROC & AUPRC & ECE \\
\midrule
Custom CNN & 0.5 & 0.697$\pm$0.032 & 0.881$\pm$0.027 & 0.512$\pm$0.076 & 0.811$\pm$0.015 & 0.748$\pm$0.042 & 0.787$\pm$0.048 & 0.178$\pm$0.044 \\
ResNet-18 & 11.2 & 0.768$\pm$0.016 & 0.967$\pm$0.019 & 0.569$\pm$0.038 & 0.870$\pm$0.009 & 0.909$\pm$0.006 & 0.930$\pm$0.004 & 0.143$\pm$0.027 \\
VGG-16 & 134.3 & 0.801$\pm$0.023 & 0.996$\pm$0.001 & 0.606$\pm$0.047 & 0.893$\pm$0.010 & 0.957$\pm$0.007 & 0.966$\pm$0.005 & 0.150$\pm$0.018 \\
ResNet-50 & 23.5 & 0.828$\pm$0.006 & 0.984$\pm$0.000 & 0.671$\pm$0.013 & 0.903$\pm$0.003 & 0.949$\pm$0.005 & 0.963$\pm$0.006 & 0.163$\pm$0.005 \\
EfficientNet-B0 & 4.0 & 0.784$\pm$0.011 & 0.963$\pm$0.007 & 0.605$\pm$0.029 & 0.876$\pm$0.004 & 0.893$\pm$0.008 & 0.921$\pm$0.010 & 0.191$\pm$0.001 \\
DenseNet-121 & 7.0 & 0.775$\pm$0.041 & 0.992$\pm$0.006 & 0.557$\pm$0.084 & 0.880$\pm$0.019 & 0.946$\pm$0.014 & 0.961$\pm$0.011 & 0.194$\pm$0.008 \\
ConvNeXt-T & 27.8 & 0.820$\pm$0.035 & 0.994$\pm$0.003 & 0.646$\pm$0.071 & 0.902$\pm$0.016 & 0.970$\pm$0.008 & 0.978$\pm$0.006 & 0.181$\pm$0.016 \\
ViT-B/16 & 85.8 & 0.811$\pm$0.022 & 0.996$\pm$0.001 & 0.626$\pm$0.043 & 0.898$\pm$0.011 & \textbf{0.974$\pm$0.008} & 0.983$\pm$0.005 & 0.143$\pm$0.048 \\
Swin-T & 27.5 & 0.809$\pm$0.016 & 0.995$\pm$0.000 & 0.623$\pm$0.031 & 0.896$\pm$0.008 & 0.969$\pm$0.005 & 0.975$\pm$0.004 & 0.177$\pm$0.028 \\
\bottomrule
\end{tabular}
\vspace{2pt}\\[-2pt]\parbox{\linewidth}{\scriptsize Bal.\,Acc.: balanced accuracy. ECE: expected calibration error (15 bins), lower is better. Sensitivity is w.r.t.\ the pneumonia class.}
\end{table}

%% file: tables/tabA_frozen_full.tex
\begin{table}[H]
\centering
\caption{Effect of unfreezing the backbone, per architecture. `Frozen' is the linear-probe protocol; `f.-t.' is the same architecture, schedule, split and augmentation with all weights trainable. Summarised in Fig.~\ref{fig:pair}b. The frozen column reproduces the linear-probe protocol used in the earlier version of this study; the fine-tuned column is the same architecture, schedule and split with all weights trainable.}
\label{tab:a-frozen}
\setlength{\tabcolsep}{2.5pt}
\scriptsize
\begin{tabular}{l rrr rrr rrr}
\toprule
Architecture & \multicolumn{3}{c}{Balanced accuracy} & \multicolumn{3}{c}{F1} & \multicolumn{3}{c}{AUROC} \\
\midrule
 & frozen & f.-t. & $\Delta$ & frozen & f.-t. & $\Delta$ & frozen & f.-t. & $\Delta$ \\
\midrule
ResNet-18 & 0.747 & 0.768$\pm$0.016 & $+$0.021 & 0.843 & 0.870$\pm$0.009 & $+$0.026 & 0.846 & 0.909$\pm$0.006 & $+$0.063 \\
VGG-16 & 0.777 & 0.801$\pm$0.023 & $+$0.024 & 0.872 & 0.893$\pm$0.010 & $+$0.021 & 0.906 & 0.957$\pm$0.007 & $+$0.052 \\
ResNet-50 & 0.782 & 0.828$\pm$0.006 & $+$0.046 & 0.869 & 0.903$\pm$0.003 & $+$0.034 & 0.909 & 0.949$\pm$0.005 & $+$0.040 \\
EfficientNet-B0 & 0.697 & 0.784$\pm$0.011 & $+$0.087 & 0.819 & 0.876$\pm$0.004 & $+$0.057 & 0.790 & 0.893$\pm$0.008 & $+$0.103 \\
DenseNet-121 & 0.835 & 0.775$\pm$0.041 & $-$0.061 & 0.903 & 0.880$\pm$0.019 & $-$0.023 & 0.946 & 0.946$\pm$0.014 & $+$0.000 \\
ConvNeXt-T & 0.827 & 0.820$\pm$0.035 & $-$0.007 & 0.900 & 0.902$\pm$0.016 & $+$0.002 & 0.938 & 0.970$\pm$0.008 & $+$0.032 \\
ViT-B/16 & 0.830 & 0.811$\pm$0.022 & $-$0.019 & 0.901 & 0.898$\pm$0.011 & $-$0.003 & 0.932 & 0.974$\pm$0.008 & $+$0.042 \\
Swin-T & 0.821 & 0.809$\pm$0.016 & $-$0.012 & 0.898 & 0.896$\pm$0.008 & $-$0.002 & 0.950 & 0.969$\pm$0.005 & $+$0.019 \\
\bottomrule
\end{tabular}
\end{table}

%% file: tables/tabA_calib_full.tex
\begin{table}[H]
\centering
\caption{Decision-threshold selection and temperature scaling, both fitted on validation data only. `Default' uses argmax at $t=0.5$ on uncalibrated soft-max scores; `tuned' uses Youden-optimal $t$ on temperature-scaled scores.}
\label{tab:a-calib-full}
\setlength{\tabcolsep}{4pt}
\footnotesize
\begin{tabular}{l rr rr rr}
\toprule
Architecture & \multicolumn{2}{c}{Balanced accuracy} & \multicolumn{2}{c}{F1} & \multicolumn{2}{c}{ECE} \\
\midrule
 & default & tuned & default & tuned & default & tuned \\
\midrule
Custom CNN & 0.652$\pm$0.052 & 0.697$\pm$0.032 & 0.809$\pm$0.013 & 0.811$\pm$0.015 & 0.139$\pm$0.079 & 0.178$\pm$0.044 \\
ResNet-18 & 0.723$\pm$0.028 & 0.768$\pm$0.016 & 0.852$\pm$0.008 & 0.870$\pm$0.009 & 0.119$\pm$0.012 & 0.143$\pm$0.027 \\
VGG-16 & 0.738$\pm$0.034 & 0.801$\pm$0.023 & 0.865$\pm$0.015 & 0.893$\pm$0.010 & 0.169$\pm$0.022 & 0.150$\pm$0.018 \\
ResNet-50 & 0.709$\pm$0.010 & 0.828$\pm$0.006 & 0.852$\pm$0.004 & 0.903$\pm$0.003 & 0.147$\pm$0.003 & 0.163$\pm$0.005 \\
EfficientNet-B0 & 0.642$\pm$0.006 & 0.784$\pm$0.011 & 0.823$\pm$0.002 & 0.876$\pm$0.004 & 0.248$\pm$0.004 & 0.191$\pm$0.001 \\
DenseNet-121 & 0.679$\pm$0.024 & 0.775$\pm$0.041 & 0.839$\pm$0.011 & 0.880$\pm$0.019 & 0.173$\pm$0.027 & 0.194$\pm$0.008 \\
ConvNeXt-T & 0.694$\pm$0.023 & 0.820$\pm$0.035 & 0.846$\pm$0.009 & 0.902$\pm$0.016 & 0.202$\pm$0.020 & 0.181$\pm$0.016 \\
ViT-B/16 & 0.742$\pm$0.064 & 0.811$\pm$0.022 & 0.866$\pm$0.028 & 0.898$\pm$0.011 & 0.150$\pm$0.053 & 0.143$\pm$0.048 \\
Swin-T & 0.699$\pm$0.045 & 0.809$\pm$0.016 & 0.848$\pm$0.019 & 0.896$\pm$0.008 & 0.193$\pm$0.037 & 0.177$\pm$0.028 \\
\bottomrule
\end{tabular}
\end{table}

%% file: tables/tabA_transfer_a.tex
\begin{table}[H]
\centering
\caption{Where the operating point is fitted, in balanced accuracy on the
official test split. `Oracle' is fitted \emph{on test} and is unattainable in
practice; it bounds the transfer gap. Prior matching needs only unlabelled
target images and an assumed prevalence, both available at deployment. Brier
decomposition and $\mathcal{A}$-distances: Table~\ref{tab:a-transfer-bc}.}
\label{tab:a-transfer}
\setlength{\tabcolsep}{8pt}
\footnotesize
\begin{tabular}{l r}
\toprule
Threshold fitted on validation (Youden)  & 0.787 \\
Threshold matched to the target prior    & \textbf{0.863} \\
Threshold fitted on test (oracle)        & 0.870 \\
\midrule
\multicolumn{2}{r}{\emph{prior matching recovers 92\% of the oracle gap}} \\
\bottomrule
\end{tabular}
\end{table}

%% file: tables/tabA_transfer_bc.tex
\begin{table}[H]
\centering
\caption{Remaining panels of Table~\ref{tab:a-transfer}.
$d_{\mathcal{A}} = 2(1-2\epsilon)$ for a domain discriminator with balanced
error $\epsilon$; $0$ means indistinguishable, $2$ perfectly separable. Panel
(b) shows temperature scaling moving reliability the \emph{wrong} way while
leaving resolution intact: the ranking survives, the probabilities do not.}
\label{tab:a-transfer-bc}
\setlength{\tabcolsep}{8pt}
\footnotesize
\begin{tabular}{l r r}
\toprule
\multicolumn{3}{l}{\emph{(b) Reliability and resolution decomposition of the Brier score}} \\
\midrule
                                  & uncalibrated & temp.-scaled \\
Reliability ($\downarrow$ better) & 0.0655 & 0.0705 \\
Resolution ($\uparrow$ better)    & 0.1078 & 0.1173 \\
\midrule
\multicolumn{3}{l}{\emph{(c) Proxy $\mathcal{A}$-distance between partitions}} \\
\midrule
Training pool vs.\ test, \textsc{normal} only    & \multicolumn{2}{r}{1.226} \\
Training pool vs.\ test, \textsc{pneumonia} only & \multicolumn{2}{r}{0.395} \\
Our train vs.\ our validation (control)          & \multicolumn{2}{r}{0.117} \\
\bottomrule
\end{tabular}
\end{table}

%% file: tables/tabA_ensemble.tex
\begin{table}[H]
\centering
\caption{Average precision on the official test split, single seed versus an
average over the three seeds already trained. Seed ensembling costs no extra
training and is selected without reference to test data. AUPRC is the
appropriate headline here because the positive class is the majority.}
\label{tab:a-ensemble}
\setlength{\tabcolsep}{5pt}
\footnotesize
\begin{tabular}{l r r r}
\toprule
Architecture & single seed & 3-seed ensemble & $\Delta$ \\
\midrule
ViT-B/16        & 0.9834 & \textbf{0.9865} & $+$0.0031 \\
Swin-T          & 0.9750 & 0.9798 & $+$0.0048 \\
VGG-16          & 0.9660 & 0.9650 & $-$0.0010 \\
ResNet-50       & 0.9599 & 0.9631 & $+$0.0032 \\
ResNet-18       & 0.9305 & 0.9448 & $+$0.0143 \\
EfficientNet-B0 & 0.9213 & 0.9344 & $+$0.0131 \\
\bottomrule
\end{tabular}
\vspace{2pt}\\[-2pt]\parbox{\linewidth}{\scriptsize
Single-seed figures are the mean over the three seeds; the ensemble averages
their temperature-scaled probabilities. The gain is largest for the weakest
backbones, which is the expected variance-reduction pattern, and is within
noise for VGG-16.}
\end{table}